\documentclass[runningheads]{llncs}
\usepackage[T1]{fontenc}
\usepackage[utf8]{inputenc}
\usepackage{graphicx}
\usepackage{hyperref}
\usepackage{booktabs}
\usepackage{xcolor}
\usepackage{multirow}
\usepackage{amssymb}

\begin{document}
\title{Knowledge Graph Entity Alignment with Graph Convolutional Networks: Lessons Learned}
\titlerunning{Knowledge Graph Entity Alignment with GCNs: Lessons Learned}
\author{
Max~Berrendorf\inst{1} \and %
Evgeniy~Faerman\inst{1} \and %
Valentyn~Melnychuk\inst{2} \and %
Volker~Tresp\inst{1,3} \and %
Thomas~Seidl\inst{1} %
}
\authorrunning{M. Berrendorf et al.}
\institute{%
Ludwig-Maximilians-Universität München, Munich, Germany\\
\email{\{berrendorf,faerman,seidl\}@dbs.ifi.lmu.de}%
\and %
Fraunhofer Institute for Integrated Circuits IIS, Germany\\
\email{v.melnychuk@campus.lmu.de}%
\and %
Siemens AG, Munich, Germany\\
\email{volker.tresp@siemens.com}%
}
\maketitle              %
\begin{abstract}
In this work, we focus on the problem of entity alignment in Knowledge Graphs (KG) and we report on our experiences when applying a Graph Convolutional Network (GCN) based model for this task.
Variants of GCN are used in multiple state-of-the-art approaches and therefore it is important to understand the specifics and limitations of GCN-based models. 
Despite serious efforts, we were not able to fully reproduce the results from the original paper and after a thorough audit of the code provided by authors, we concluded, that their implementation is different from the architecture described in the paper.
In addition, several tricks are required to make the model work and some of them are not very intuitive. 
We provide an extensive ablation study to quantify the effects these tricks and changes of architecture have on final performance.
Furthermore, we examine current evaluation approaches and systematize available benchmark datasets.
We believe that people interested in KG matching might profit from our work, as well as novices entering the field.\footnote{Code: \href{https://github.com/Valentyn1997/kg-alignment-lessons-learned}{https://github.com/Valentyn1997/kg-alignment-lessons-learned}.}%

\end{abstract}
\section{Introduction}
The success of information retrieval in a given task critically depends on the quality of the underlying data. Another issue is that in many domains knowledge bases are spread across various data sources \cite{nickel2015review} and it is crucial to be able to combine information from different sources.
In this work, we focus on knowledge bases in the form of Knowledge Graphs (KGs), which are particularly suited for information retrieval \cite{singhal2012introducing}.
Joining information from different KGs is non-trivial, as there is no unified schema or vocabulary. The goal of the entity alignment task is to overcome this problem by \emph{learning} a matching between entities in different KGs. In the typical setting some of the alignments are known in advance (seed alignments) and the task is therefore supervised. More formally, we are given graphs $G_L = (V_L, E_L)$ and $G_R = (V_R, E_R)$ 
with a seed alignment $A = {(l_i, r_i)}_i \subseteq V_L \times V_R$.
It is commonly assumed that an entity $v \in V_L$ can match at most one entity $v' \in V_R$.
Thus the goal is to infer alignments for the remaining nodes only.

Graph Convolutional Networks(GCN) \cite{kipf2016semi,gilmer2017neural}, which have been recently become increasingly popular, are at the core of state-of-the-art methods for entity alignments in KGs \cite{DBLP:conf/emnlp/WangLLZ18,DBLP:conf/acl/CaoLLLLC19,xu2019cross,DBLP:conf/ijcai/ZhuZ0TG19,anonymous2020deep}.
In this paper, we thoroughly analyze one of the first GCN-based entity alignment methods, GCN-Align~\cite{DBLP:conf/emnlp/WangLLZ18}.
Since the other methods we are studying can be considered as extensions of this first paper and have a similar architecture, our goal is to understand the importance of its individual components and architecture choices.
\noindent In summary, our contribution is as follows:
\begin{enumerate}
    \item We investigate the reproducibility of the published results of a recent GCN-based method for entity alignment and uncover differences between the method's description in the paper and the authors' implementation.
    \item We perform an ablation study to demonstrate the individual components' contribution.
    \item We apply the method to numerous additional datasets of different sizes to investigate the consistency of results across datasets.
\end{enumerate}

\section{Related work}
In this section we review previous work for the entity alignment for Knowledge Graphs and revise datasets and current evaluation process. We believe this is useful for practitioners, since we discover some pitfalls, especially when implementing evaluation scores and selecting datasets for comparison. The overview of methods, datasets and metrics is provided in Table~\ref{tab:related}.
\begin{table}
    \centering
    \caption{Overview of related work in the field of entity alignment for knowledge graphs with their used datasets and metrics.}
    \label{tab:related}
    \scriptsize
\begin{tabular*}{\linewidth}{*{4}{@{\extracolsep{\fill}}l}@{\extracolsep{\fill}}}
\toprule
Method & Datasets & Metrics & Code\\ %
\midrule
MTransE \cite{DBLP:conf/ijcai/ChenTYZ17} & WK3l-15K, WK3l-120K, CN3l & H@10(, MR) & yes\\%
IPTransE \cite{DBLP:conf/ijcai/ZhuXLS17} & DFB-\{1,2,3\} & H@\{1,10\}, MR & yes\\%
JAPE \cite{DBLP:conf/semweb/SunHL17} & DBP15K(JAPE) & H@\{1,10,50\}, MR & yes\\%
KDCoE \cite{DBLP:conf/ijcai/ChenTCSZ18} & WK3l-60K & H@\{1,10\}, MR & yes\\%
BootEA \cite{DBLP:conf/ijcai/SunHZQ18} & DBP15K(JAPE), DWY100K & H@\{1,10\}, MRR & yes\\%
SEA \cite{DBLP:conf/www/PeiYHZ19} & WK3l-15K, WK3l-120K & H@\{1,5,10\}, MRR & yes\\%
MultiKE \cite{DBLP:conf/ijcai/ZhangSHCGQ19} & DWY100K & H@\{1,10\}, MR, MRR & yes\\%
AttrE \cite{DBLP:conf/aaai/TrisedyaQZ19} & DBP-LGD,DBP-GEO,DBP-YAGO & H@\{1,10\}, MR & yes\\%
RSN \cite{DBLP:conf/icml/GuoSH19} & custom DBP15K, DWY100K & H@\{1,10\}, MRR &  yes\\%
\textbf{GCN-Align} \cite{DBLP:conf/emnlp/WangLLZ18} & DBP15K(JAPE) & H@\{1,10,50\}& yes\\%
CL-GNN \cite{DBLP:conf/acl/XuWYFSWY19} & DBP15K(JAPE) & H@\{1,10\} & yes\\%
MuGNN \cite{DBLP:conf/acl/CaoLLLLC19} & DBP15K(JAPE), DWY100K & H@\{1,10\}, MRR & yes\\%
NAEA \cite{DBLP:conf/ijcai/ZhuZ0TG19} & DBP15K(JAPE), DWY100K & H@\{1,10\}, MRR & no \\
\bottomrule
\end{tabular*}    
 \end{table}
\subsection{Methods}
While the problem of entity alignments in Knowledge Graphs has been tackled historically by researching vocabularies which are as broad as possible, and establish them as a standard, recent approaches take a more data-driven view.
Early methods use classical knowledge graph link prediction models such as TransE~\cite{DBLP:conf/nips/BordesUGWY13} to embed the entities of the individual knowledge graphs using a intra-KG link prediction loss, and differ in what they do with the aligned entities.
For instance MTransE~\cite{DBLP:conf/ijcai/ChenTYZ17} learns a linear transformation between the embedding spaces of the individual graphs using $L_2$-loss.
BootEA~\cite{sun2018bootstrapping} adopts a bootstrapping approach and iteratively labels the most likely alignments and utilizes them for further training. In addition to the alignment loss, embeddings of aligned entities are swapped regularly to calibrate embedding spaces against each other. SEA~\cite{pei2019semi} learns mapping between embedding spaces in both directions and additionaly adds cycle-consistency loss. Therefore the distance between original embedding of an entity and its representation, which was first translated to another space and then back from it, is penalized.
IPTransE~\cite{DBLP:conf/ijcai/ZhuXLS17} embeds both KGs into the same embedding space and uses a margin-based loss to enforce the embeddings of aligned entities to become similar.
RSN~\cite{DBLP:conf/icml/GuoSH19} model generates sequences using different types of random walks which can move between graphs when visiting aligned entities. The generated sequences are feed to adapted recurrent model.
JAPE~\cite{DBLP:conf/semweb/SunHL17}, KDCoE~\cite{DBLP:conf/ijcai/ChenTCSZ18}, MultiKE~\cite{DBLP:conf/ijcai/ZhangSHCGQ19} and AttrE \cite{DBLP:conf/aaai/TrisedyaQZ19} utilize attributes available for some entities and additional information like names of entities and relationships.
Graph Neural Network (GNN) based models
\cite{DBLP:conf/emnlp/WangLLZ18,DBLP:conf/acl/CaoLLLLC19,xu2019cross,DBLP:conf/ijcai/ZhuZ0TG19,anonymous2020deep}\footnote{Please note, that while \cite{DBLP:conf/ijcai/ZhuZ0TG19} does not state explicitly that they use GNNs, their model is very similar to \cite{velivckovic2017graph}.} have in common that they use GNN to create node representations by aggregating node representations together with representations of their neighbors.
Most of GNN approaches do not distinguish between different relations and either consider all neighbors equally \cite{DBLP:conf/emnlp/WangLLZ18,xu2019cross,anonymous2020deep} or use attention \cite{DBLP:conf/acl/CaoLLLLC19} to weight the representations of the neighbors for the aggregation.

\subsection{Datasets}
\begin{table}[ht]
    \centering
    \caption{
    Overview of used datasets with their sizes in the number of triples (edges), entities (nodes), relations (different edge types) and alignments.
    For Wk3l, the alignment is provided as a directed mapping on a entity level.
    However, there are additional triple alignments.
    Following a common practice as e.g. \cite{DBLP:conf/www/PeiYHZ19} we can assume that an alignment should be symmetric, and that we can extract entity alignments from the triple alignments.
    Doing so, we obtain the number of alignments given in brackets.
    }
    \label{tab:my_label}
    \scriptsize
    \begin{tabular*}{\linewidth}{*{2}{l}*{5}{@{\extracolsep{\fill}}r}@{\extracolsep{\fill}}}
    \toprule
    Dataset & Subset & Graph & Triples & Entities & Relations & Alignments \\
    \midrule
    \multirow{6}{*}{DBP15k (full)}
    & \multirow{2}{*}{fr-en} & fr & 
    192,191 & 66,858 & 1,379 & \multirow{2}{*}{15,000}\\
    & & en & 
    278,590 & 105,889 & 2,209 & \\
    & \multirow{2}{*}{ja-en} & ja &
    164,373 & 65,744 & 2,043 & \multirow{2}{*}{15,000}\\
    & & en & 
    233,319 & 95,680 & 2,096 & \\
    & \multirow{2}{*}{zh-en} & zh &
    153,929 & 66,469 & 2,830 & \multirow{2}{*}{15,000}\\
    & & en & 
    237,674 & 98,125 & 2,317 & \\
    \midrule
    \multirow{6}{*}{DBP15k (JAPE)}
    & \multirow{2}{*}{fr-en} & fr & 
    105,998 & 19,661 & 903 & \multirow{2}{*}{15,000}\\
    & & en & 
    115,722 & 19,993 & 1,208 & \\
    & \multirow{2}{*}{ja-en} & ja &
    77,214 & 19,814 & 1,299 & \multirow{2}{*}{15,000}\\
    & & en & 
    93,484 & 19,780 & 1,153 & \\
    & \multirow{2}{*}{zh-en} & zh &
    70,414 & 19,388 & 1,701 & \multirow{2}{*}{15,000}\\
    & & en & 
    95,142 & 19,572 & 1,323 & \\
    \midrule
    \multirow{4}{*}{WK3l-15k}
    & \multirow{2}{*}{en-de} & en &
    209,041 & 15,127 & 1,841 & 1,289 (10,383) \\
    & & de & 
    144,244 & 14,603 & 596 & 1,140 (10,383) \\
    & \multirow{2}{*}{en-fr} & en &
    203,356 & 15,170 & 2,228 & 2,498 (\phantom{1}8,024) \\
    & & fr & 
    169,329 & 15,393 & 2,422 & 3,812 (\phantom{1}8,024) \\
    \midrule
    \multirow{4}{*}{WK3l-120k}
    & \multirow{2}{*}{en-de} & en &
    624,659 & 67,650 & 2,393 & 6,173 (50,280) \\
    & & de & 
    389,554 & 61,942 & 861 & 4,820 (50,280) \\
    & \multirow{2}{*}{en-fr} & en &
    1,375,406 & 119,749 & 3,109 & 36,749 (87,836) \\
    & & fr & 
    760,497 & 118,592 & 2,336 & 36,013 (87,836) \\
    \midrule
    \multirow{4}{*}{DWY-100k}
    & \multirow{2}{*}{dbp-wd} & dbp &
    463,294 & 100,000 & 330 & \multirow{2}{*}{100,000}\\
    & & wd & 
    448,774 & 100,000 & 220 & \\
    & \multirow{2}{*}{dbp-yg} & dbp &
    428,952 & 100,000 & 302 & \multirow{2}{*}{100,000}\\
    & & yg & 
    502,563 & 100,000 & 31 & \\
    \bottomrule
    \end{tabular*}
\end{table}
The datasets used by entity alignments methods generally derive from large-scale open-source data source such as DBPedia~\cite{DBLP:conf/semweb/AuerBKLCI07}, YAGO~\cite{DBLP:conf/cidr/MahdisoltaniBS15}, or Wikidata~\cite{wikidata}.
While there is the DWY-100k dataset, which comprises 100k aligned entities across the three aforementioned individual knowledge graphs, most of the datasets, such as DBP15k, or WK3l derive from a single multi-lingual database.
There, subsets are formed according to a specific language, and entities which occur in multiple languages and are linked accordingly are used as alignments.

As an interesting observation we found out that all papers which evaluate on DBP15k, do not evaluate on the full DBP15k dataset\footnote{Available at \url{http://ws.nju.edu.cn/jape/}} (which we refer to as \emph{DBP15k (full)}), but rather use a smaller subset provided by the authors of JAPE~\cite{DBLP:conf/semweb/SunHL17} in their GitHub repository\footnote{\url{https://github.com/nju-websoft/JAPE/blob/master/data/dbp15k.tar.gz}}, which we call \emph{DBP15k-JAPE}.
The smaller subsets were created by selecting a portion of entities (around 20k of 100k) which are popular, i.e. appear in many triples as head or tail.
The number of aligned entities stays the same (15k).
As the paper only reports the dataset statistics of the larger dataset, and does not mention the reduction of the dataset, subsequent papers
also report the statistics of the larger dataset, although experiments use the smaller variant \cite{DBLP:conf/semweb/SunHL17,sun2018bootstrapping,DBLP:conf/emnlp/WangLLZ18,DBLP:conf/acl/CaoLLLLC19,DBLP:conf/ijcai/ZhuXLS17}.

\subsection{Scores}
It is common practice to only consider the entities being part of the test alignment as potential matching candidates.
Although we argue that ignoring entities exclusive to a single graph as potential candidates does not reflect well the use-case situation\footnote{In the typical scenario it is not known in advance, which entities have matching and which not. Therefore the resulting score is too optimistic. However, we advocate to investigate this shortcoming further in future work}, we follow this evaluation scheme for our experiments to maintain comparability.

In the following description of evaluation measures we focus only on the case of aligning one node $l_i \in V_L$ with a ground truth alignment $r_i \in V_R$.
The right-to-left alignment is handled analogously.
Let $V_R^* = \{v_r \in V_R \mid \exists v_l \in V_L: (v_l, v_r) \in A_e\}$ denote the set of matching candidates in the right graph.
For a node $l_i$, the entity alignment models generates a score $f(l_i, v_j)$ for each matching candidate $v_j \in V_R^*$.
Afterwards, the candidates are sorted according to their score, and the rank $rank(l_i, r_i)$ is computed as the index of the ground truth match $r_i$ in this sorted list (1-based).
The \emph{mean rank (MR)} is simply the mean over the ranks for all alignments.
$$
MR = \frac{1}{|A_e|}\sum \limits_{i=1}^{|A_e|} rank(l_i, r_i)
$$
The \emph{mean reciprocal rank (MRR)} is the mean over all reciprocal ranks.
$$
MRR = \frac{1}{|A_e|}\sum \limits_{i=1}^{|A_e|} \frac{1}{rank(l_i, r_i)}
$$
It is naturally bounded between 0 and 1, where 1 corresponds to a perfect score.
Moreover, its value is dominated by small ranks, and it is less sensitive to larger ones.
The \emph{hits at $k$ (H@k)} is the percentage of alignments where the rank was at most $k$, i.e. equivalent to the recall at $k$.
$$
H@k = \frac{|\{(l_i, r_i) \in A_e \mid rank(l_i, r_i) \leq k\}|}{|A_e|}
$$ %
\section{Method}
GCN-Align~\cite{DBLP:conf/emnlp/WangLLZ18} is a GCN-based approach to embed all entities from both graphs into a common embedding space.
Each entity $i$ is associated with \emph{structural} features $h_i^s \in \mathbb{R}^d$, which are initialized randomly and updated during training.
The features of all entities in a single graph are combined to the feature matrix $H^s$.
Subsequently, a two-layer GCN is applied.
A single GCN layer is described by
$$
H^{(i+1)} = \sigma\left(\hat{D}^{-\frac{1}{2}}\hat{A}\hat{D}^{-\frac{1}{2}}H^{(i)}W^{(i)}\right)
$$
with $\hat{A} = A + I$, where $A$ is the adjacency matrix, and $\hat{D}_{ii} = \sum \limits_{j=1}^n \hat{A}_{ij}$ is the diagonal node degree matrix.
The input of the first layer is set to $H^{(0)} = H^s$, and $\sigma$ is non-linear activation function.
For the first layer, $\sigma=ReLU$, and the second layer uses the identity.
The output of the second layer is considered as the structural embedding, denoted by $s_i = H^{(2)}_i \in \mathbb{R}^{d}$.
Both graphs are equipped with its own node features, but the convolution weights $W^{(i)}$ are shared across the graphs.

The adjacency matrix is derived from the knowledge graph by first computing a score, called \emph{functionality}, for each relation as the ratio between the number of different entities which occur as head, and the number of triples in which the relation occurs
$$
\alpha_r := \frac{|\{v \in V \mid \exists v': (v, r, v') \in T\}|}{|\{t \in T \mid \exists v, v' \in V: (v, r, v) \in T\}|}
$$
Analogously, the \emph{inverse functionality} $\alpha_r'$ is obtained by replacing the nominator by the number of different tail entities.
The final adjacency matrix is obtained as
$$
A_ij = \sum \limits_{(e_i, r, e_j)} \alpha_r' + \sum \limits_{(e_j, r, e_j)} \alpha_r
$$

In addition, each entity $i$ is also equipped with attributes $h_i^a \in \mathbb{R}^{d'}$ which are combined into a graph attribute matrix $H^a$.
Anagously, the attributes are processed by a GCN for each graph with convolution weights shared across the graphs, resulting in attribute embeddings $a_i \in \mathbb{R}^{d'}$.

The attribute and structure GCNs are optimized separately using SGD.
As loss function, a margin-rank loss is used, exemplary for the structure embedding
$$
L = \sum \limits_{(r_i, l_i) \in A} \sum \limits_{(r_j, l_j) \in A_i^-} \left[\|s_i^L - s_i^R\|_1 + \gamma - \|s_j^L - s_j^R\|\right]_{+}
$$
where $[x]_+ = \max \{0, x\}$, and the margin $\gamma$ is a hyperparameter chosen separately for structure and attribute embeddings.
$A_i^-$ denotes a set of negative samples constructed by either replacing the left or the right entity with a random entity from the same graph.

In order to compare two nodes from both graphs, the $L_1$ distance between their embeddings is used, normalised by the dimensionality.
$$
score(v_i^L, v_j^R) = - \left(\beta \frac{\|s_i^L - s_j^R\|_1}{d} + (1 - \beta) \frac{\|a_i^L - a_j^R\|_1}{d'} \right)
$$
Here, $\beta$ is a hyperparameter for the tradeoff between structural and attribute similarity.

\subsection{Implementation Differences}
The code\footnote{\url{https://github.com/1049451037/GCN-Align}} provided by the authors differs in a few aspects from the method described in the paper.
First, instead of using the full DBP15k dataset having the dataset sizes as reported in the paper, a smaller version is used.
Second, when computing the adjacency matrix, $fun(r)$ and $ifun(f)$ are set to at least 0.3.
Third, the node features are always normalised to unit Euclidean length before passing them into the network.
Finally, there are no convolution weights.
This fact is particularly interesting, as this means that the whole GCN does not contain a single parameter, but is just a fixed function on the learned node embeddings. %

\section{Experiments}
In initial experiments we were able to reproduce the results reported in the paper using the implementation provided by the authors.
Moreover, we are able to reproduce the results using our own implementation, and settings adjusted to the authors' code.
In addition, we replaced the adjacency matrix based on functionality and inverse functionality by a simpler version, where $a_{ij} = \{(h,r,t) \in T \mid h=e_i, t=e_j\}$.
We additionally use $\hat{D}^{-1}\hat{A}$ instead of the symmetric normalization.
In total, we see no difference in performance between our simplified adjacency matrix, and the authors' one.
We identified two aspects which affect the model's performance:
Not using convolutional weights, and normalizing the variance when initializing node embeddings.
We provide empirical evidence for this finding across numerous datasets.
\begin{table}
    \centering
    \caption{Hyperparameter grid used for large-scale hyperparameter search on \emph{DBP15K (JAPE) zh-en}.}
    \begin{tabular*}{\linewidth}{l@{\extracolsep{\fill}}l@{\extracolsep{\fill}}r}
        \toprule
        hyperparameter & abbrev. & value range\\
        \midrule
        optimizer & opt & \{Adam, SGD\}\\
        learning rate & lr & \{0.1, 0.5, 1, 10, 20\}\\
        number of layers & \#layers & \{1, 2, 3\}\\
        number of negative samples & \#neg & \{5, 50, 100\}\\
        number of epochs & \#epochs & \{10, 500, 2000, 3000\}\\
        \bottomrule
    \end{tabular*}
    \label{tab:hyperparameters}
\end{table}

\begin{table}
    \centering
    \caption{Optimal Hyperparameters found for DBP15k (JAPE), zh-en with and without convolution weights, and with two different embedding initialization variances.}
    \begin{tabular*}{\linewidth}{l*{4}{@{\extracolsep{\fill}}r}}
        \toprule
        \textbf{Weights} & \multicolumn{2}{c}{no} & \multicolumn{2}{c}{yes} \\
        \textbf{Variance Emb. Init.} &  1 & $n^{-1/2}$ &  1 & $n^{-1/2}$ \\
        \midrule
        \#epochs    &  2,000 &  3,000 &  2,000 &  2,000 \\
        \#neg &    50 &   100 &    50 &    50 \\
        \#layers      &     2 &     2 &     3 &     2 \\
        lr            &     1 &     1 &     1 &     1 \\
        opt     &  adam &   sgd &  adam &  adam \\
        \bottomrule
    \end{tabular*}
    \label{tab:best_zh_en}
\end{table}

\begin{table}
    \centering
    \caption{
    Ablation study on using convolution weights and different embedding initialisation.
    A detailed description can be found in the text.
    }
    \label{tab:result}
    \scriptsize
\begin{tabular*}{\linewidth}{*{2}{l}*{10}{@{\extracolsep{\fill}}r}@{\extracolsep{\fill}}}
\toprule
        & \textbf{Weights} & \multicolumn{2}{c}{No} & \multicolumn{2}{c}{Yes}\\
         \multicolumn{2}{r}{\textbf{Variance. Emb. Init.}} & $1$ & $n^{-1/2}$ & $1$ &  $n^{-1/2}$ \\
         \midrule
         & {} & \multicolumn{4}{c}{H@1} \\
        \midrule
        \textbf{DBP15K (full)} & \textbf{fr-en} &  31.51 $\pm$ 0.16 &  27.64 $\pm$ 0.22 &  21.82 $\pm$ \phantom{0}0.39 &  16.73 $\pm$ \phantom{0}0.59\\
        & \textbf{ja-en} &  33.26 $\pm$ 0.10 &  29.06 $\pm$ 0.23 &  26.21 $\pm$ \phantom{0}0.33 &  20.78 $\pm$ \phantom{0}0.16 \\
        & \textbf{zh-en} &  31.15 $\pm$ 0.15 &  22.55 $\pm$ 0.27 &  24.96 $\pm$ \phantom{0}0.71 &  18.85 $\pm$ \phantom{0}0.99 \\
        \midrule
\textbf{DBP15K (JAPE)} & \textbf{fr-en} &  45.37 $\pm$ 0.13 &  41.03 $\pm$ 0.13 &  35.36 $\pm$ \phantom{0}0.33 &  30.50 $\pm$ \phantom{0}0.38 \\
        & \textbf{ja-en} &  45.53 $\pm$ 0.18 &  40.29 $\pm$ 0.09 &  35.81 $\pm$ \phantom{0}0.53 &  31.46 $\pm$ \phantom{0}0.15 \\
        & \textbf{zh-en} &  43.30 $\pm$ 0.12 &  39.37 $\pm$ 0.20 &  33.61 $\pm$ \phantom{0}0.49 &  29.94 $\pm$ \phantom{0}0.35 \\
        \midrule
\textbf{DWY100K} & \textbf{wd} &  58.50 $\pm$ 0.05 &  54.07 $\pm$ 0.05 &  50.13 $\pm$ \phantom{0}0.11 &  38.85 $\pm$ \phantom{0}0.31 \\
        & \textbf{yg} &  72.82 $\pm$ 0.06 &  67.06 $\pm$ 0.03 &  67.36 $\pm$ \phantom{0}0.10 &  60.67 $\pm$ \phantom{0}0.30 \\
        \midrule
\textbf{WK3l-120K} & \textbf{en-de} &  10.10 $\pm$ 0.03 &   9.17 $\pm$ 0.05 &   9.02 $\pm$ \phantom{0}0.17 &   6.75 $\pm$ \phantom{0}0.12 \\
        & \textbf{en-fr} &   8.28 $\pm$ 0.03 &   7.38 $\pm$ 0.03 &   7.26 $\pm$ \phantom{0}0.11 &   5.07 $\pm$ \phantom{0}0.16 \\
        \midrule
\textbf{WK3l-15K} & \textbf{en-de} &  16.57 $\pm$ 0.12 &  14.41 $\pm$ 0.23 &  17.43 $\pm$ \phantom{0}0.38 &  12.66 $\pm$ \phantom{0}0.30 \\
        & \textbf{en-fr} &  17.07 $\pm$ 0.15 &  16.16 $\pm$ 0.16 &  15.98 $\pm$ \phantom{0}0.16 &  12.41 $\pm$ \phantom{0}0.18 \\
 \midrule       
     & {} & \multicolumn{4}{c}{H@10} \\
\midrule
\textbf{DBP15K (full)} & \textbf{fr-en} &  68.38 $\pm$ 0.32 &  63.41 $\pm$ \phantom{0}0.14 &  59.26 $\pm$ \phantom{0}0.55 &  48.55 $\pm$ \phantom{0}0.92 \\
        & \textbf{ja-en} &  68.22 $\pm$ 0.09 &  61.95 $\pm$ 0.17 &  61.12 $\pm$ \phantom{0}0.51 &  49.56 $\pm$ \phantom{0}0.38 \\
        & \textbf{zh-en} &  67.46 $\pm$ 0.11 &  56.03 $\pm$ 0.21 &  59.07 $\pm$ \phantom{0}1.10 &  50.32 $\pm$ \phantom{0}1.52 \\
        \midrule
\textbf{DBP15K (JAPE)} & \textbf{fr-en} &  82.48 $\pm$ 0.08 &  79.11 $\pm$ 0.07 &  74.71 $\pm$ \phantom{0}0.27 &  69.72 $\pm$ \phantom{0}0.36 \\
        & \textbf{ja-en} &  79.77 $\pm$ 0.14 &  75.13 $\pm$ 0.20 &  73.05 $\pm$ \phantom{0}0.52 &  67.18 $\pm$ \phantom{0}0.28 \\
        & \textbf{zh-en} &  77.63 $\pm$ 0.05 &  73.66 $\pm$ 0.28 &  71.16 $\pm$ \phantom{0}0.17 &  66.22 $\pm$ \phantom{0}0.51 \\
        \midrule
\textbf{DWY100K} & \textbf{wd} &  86.26 $\pm$ 0.05 &  81.30 $\pm$ 0.03 &  79.65 $\pm$ \phantom{0}0.20 &  69.73 $\pm$ \phantom{0}0.25 \\
        & \textbf{yg} &  92.13 $\pm$ 0.04 &  87.57 $\pm$ 0.04 &  88.64 $\pm$ \phantom{0}0.09 &  83.76 $\pm$ \phantom{0}0.27 \\
        \midrule
\textbf{WK3l-120K} & \textbf{en-de} &  27.13 $\pm$ 0.02 &  24.92 $\pm$ 0.03 &  25.49 $\pm$ \phantom{0}0.26 &  20.83 $\pm$ \phantom{0}0.29 \\
        & \textbf{en-fr} &  23.73 $\pm$ 0.04 &  21.57 $\pm$ 0.05 &  22.16 $\pm$ \phantom{0}0.24 &  16.31 $\pm$ \phantom{0}0.35 \\
        \midrule
\textbf{WK3l-15K} & \textbf{en-de} &  42.43 $\pm$ 0.13 &  38.63 $\pm$ 0.17 &  45.24 $\pm$ \phantom{0}0.47 &  37.03 $\pm$ \phantom{0}0.30 \\
        & \textbf{en-fr} &  49.68 $\pm$ 0.15 &  48.18 $\pm$ 0.15 &  47.64 $\pm$ \phantom{0}0.53 &  41.87 $\pm$ \phantom{0}0.36 \\
\midrule
        & {} & \multicolumn{4}{c}{MR}\\
        \midrule
        \textbf{DBP15K (FULL)} & \textbf{fr-en} &   203.90 $\pm$ 3.80 &   262.24 $\pm$ 3.23 &   123.09 $\pm$ 15.43 &   208.00 $\pm$ 12.04 \\
        & \textbf{ja-en} &   206.17 $\pm$ 4.21 &   358.53 $\pm$ 3.65 &   138.80 $\pm$ 12.87 &   238.24 $\pm$ 24.09 \\
        & \textbf{zh-en} &   168.80 $\pm$ 2.59 &   149.08 $\pm$ 2.70 &   279.49 $\pm$ 38.78 &   206.36 $\pm$ 17.60 \\
        \midrule
\textbf{DBP15K (JAPE)} & \textbf{fr-en} &   109.64 $\pm$ 1.56 &   117.59 $\pm$ 2.91 &    130.75 $\pm$ \phantom{0}8.48 &    133.14 $\pm$ \phantom{0}7.09 \\
        & \textbf{ja-en} &   144.81 $\pm$ 1.89 &   195.19 $\pm$ 3.44 &    146.42 $\pm$ \phantom{0}6.45 &   221.92 $\pm$ 12.22 \\
        & \textbf{zh-en} &   181.37 $\pm$ 4.05 &   215.23 $\pm$ 4.53 &   172.05 $\pm$ 12.72 &    236.72 $\pm$ \phantom{0}2.84 \\
         \midrule
\textbf{DWY100K} & \textbf{wd} &   277.08 $\pm$ 8.28 &   460.32 $\pm$ 9.17 &   500.61 $\pm$ 24.10 &   563.29 $\pm$ 28.92 \\

        & \textbf{yg} &    49.32 $\pm$ 2.71 &   102.50 $\pm$ 3.69 &    105.52 $\pm$ \phantom{0}4.63 &     67.71 $\pm$ \phantom{0}3.82 \\
         \midrule
\textbf{WK3l-120K} & \textbf{en-de} &  2753.75 $\pm$ 6.69 &  2280.31 $\pm$ 8.97 &  2843.96 $\pm$ 53.29 &  2289.02 $\pm$ 36.71 \\
        & \textbf{en-fr} &  4438.81 $\pm$ 9.29 &  4110.23 $\pm$ 7.90 &  4551.39 $\pm$ 55.29 &  4007.91 $\pm$ 59.01 \\
         \midrule
\textbf{WK3l-15K} & \textbf{en-de} &   247.74 $\pm$ 1.09 &   233.29 $\pm$ 2.66 &    263.16 $\pm$ \phantom{0}6.75 &    197.40 $\pm$ \phantom{0}5.39 \\
        & \textbf{en-fr} &   196.16 $\pm$ 1.09 &   176.32 $\pm$ 1.03 &    249.77 $\pm$ \phantom{0}7.71 &    184.72 $\pm$ \phantom{0}3.32 \\
\midrule
        & {} & \multicolumn{4}{c}{MRR}\\
        \midrule
        \textbf{DBP15K (full)} & \textbf{fr-en} &  43.59 $\pm$ 0.08 &  39.30 $\pm$ 0.18 &  33.83 $\pm$ \phantom{0}0.46 &  27.03 $\pm$ \phantom{0}0.72 \\
        & \textbf{ja-en} &  44.68 $\pm$ 0.06 &  39.92 $\pm$ 0.20 &  37.59 $\pm$ \phantom{0}0.37 &  30.31 $\pm$ \phantom{0}0.19 \\
        & \textbf{zh-en} &  43.09 $\pm$ 0.10 &  33.55 $\pm$ 0.19 &  36.15 $\pm$ \phantom{0}0.80 &  29.21 $\pm$ \phantom{0}1.16 \\
        \midrule
\textbf{DBP15K (JAPE)} & \textbf{fr-en} &  57.95 $\pm$ 0.10 &  53.78 $\pm$ 0.05 &  48.31 $\pm$ \phantom{0}0.26 &  43.37 $\pm$ \phantom{0}0.33 \\
        & \textbf{ja-en} &  57.14 $\pm$ 0.13 &  51.96 $\pm$ 0.07 &  48.03 $\pm$ \phantom{0}0.55 &  43.30 $\pm$ \phantom{0}0.15 \\
        & \textbf{zh-en} &  54.89 $\pm$ 0.09 &  50.88 $\pm$ 0.15 &  45.97 $\pm$ \phantom{0}0.39 &  41.93 $\pm$ \phantom{0}0.40 \\
        \midrule
\textbf{DWY100K} & \textbf{wd} &  68.33 $\pm$ 0.03 &  63.68 $\pm$ 0.04 &  60.50 $\pm$ \phantom{0}0.14 &  49.56 $\pm$ \phantom{0}0.30 \\
        & \textbf{yg} &  79.74 $\pm$ 0.04 &  74.29 $\pm$ 0.03 &  74.93 $\pm$ \phantom{0}0.09 &  68.76 $\pm$ \phantom{0}0.28 \\
        \midrule
\textbf{WK3l-120K} & \textbf{en-de} &  16.05 $\pm$ 0.03 &  14.73 $\pm$ 0.03 &  14.73 $\pm$ \phantom{0}0.21 &  11.70 $\pm$ \phantom{0}0.18 \\
        & \textbf{en-fr} &  13.65 $\pm$ 0.02 &  12.34 $\pm$ 0.02 &  12.41 $\pm$ \phantom{0}0.16 &   9.04 $\pm$ \phantom{0}0.23 \\
        \midrule
\textbf{WK3l-15K} & \textbf{en-de} &  25.40 $\pm$ 0.10 &  22.81 $\pm$ 0.16 &  26.94 $\pm$ \phantom{0}0.38 &  21.06 $\pm$ \phantom{0}0.25 \\
        & \textbf{en-fr} &  27.98 $\pm$ 0.16 &  26.76 $\pm$ 0.13 &  26.55 $\pm$ \phantom{0}0.23 &  22.20 $\pm$ \phantom{0}0.21 \\
\bottomrule
\end{tabular*}
\end{table}
We fix using convolution weights and the variance for the normal distribution from which the embedding vectors are initialized and optimize the other hyperparameters according to validation H@1 (80/20\% train-validation split) on \emph{DBP15K (JAPE) zh-en} in a large-scale hyperparameter search, comprising 1,440 experiments.
The hyperparameter grid is given in Table~\ref{tab:hyperparameters}, and Table~\ref{tab:best_zh_en} shows the best parameters found for DBP15k (JAPE) zh-en for the four different settings.
For each dataset, we perform a smaller hyperparameter search to fine-tune LR, \#epochs \& \#layers for each dataset (again 80/20 split).
Their optimal parameters are given in the appendix, in Table~\ref{tab:best_hparams}.
We evaluate the best models on the official test set.
Our results regarding Hits@1 (H@1), Hits@10 (H@10), mean rank (MR) and mean reciprocal rank (MRR) are summarised in Table~\ref{tab:result}.

\paragraph{Node Embedding Initialization}
Comparing the columns of Table~\ref{tab:result} we can observe the influence of the node embedding initialization.
Using the settings from the authors' code, i.e. not using weights, a choosing a variance of $n^{-1/2}$ actually results in inferior performance in terms of H@1, as compared to use a standard normal distribution.
These findings are consistent across datasets.

\paragraph{Convolution Weights}
The first column of Table~\ref{tab:result} corresponds to the weight usage and initialization settings used in the code for GCN-Align.
We achieve slightly better results than published in \cite{DBLP:conf/emnlp/WangLLZ18}, which we attribute to a more exhaustive parameter search.
Interestingly, all best configurations use Adam optimizer instead of SGD.
Adding convolution weights degrades the performance across all datasets and subsets thereof but one as witnessed by comparing the first two columns with the last two columns.

\section{Conclusion}
In this work, we reported our experiences when implementing the Knowledge Graph alignment method GCN-Align. We pointed at important differences between the model described in the paper and the actual implementation and quantified their effects in the ablation study. 
For future work, we plan to include other methods for entity alignments in our framework.
\section*{Acknowledgements}
This work has been funded by the German Federal Ministry of Education and Research (BMBF) under Grant No. 01IS18036A and by the Bavarian Ministry for Economic Affairs, Infrastructure, Transport and Technology through the Center for Analytics-Data-Applications (ADA-Center) within the framework of “BAYERN DIGITAL II”. The authors of this work take full responsibilities for its content.

\bibliographystyle{splncs04}
\bibliography{main}

\begin{thebibliography}{10}
\providecommand{\url}[1]{\texttt{#1}}
\providecommand{\urlprefix}{URL }
\providecommand{\doi}[1]{https://doi.org/#1}

\bibitem{DBLP:conf/semweb/AuerBKLCI07}
Auer, S., Bizer, C., Kobilarov, G., Lehmann, J., Cyganiak, R., Ives, Z.G.:
  Dbpedia: {A} nucleus for a web of open data. In: Aberer, K., Choi, K., Noy,
  N.F., Allemang, D., Lee, K., Nixon, L.J.B., Golbeck, J., Mika, P., Maynard,
  D., Mizoguchi, R., Schreiber, G., Cudr{\'{e}}{-}Mauroux, P. (eds.) The
  Semantic Web, 6th International Semantic Web Conference, 2nd Asian Semantic
  Web Conference, {ISWC} 2007 + {ASWC} 2007, Busan, Korea, November 11-15,
  2007. Lecture Notes in Computer Science, vol.~4825, pp. 722--735. Springer
  (2007). \doi{10.1007/978-3-540-76298-0\_52},
  \url{https://doi.org/10.1007/978-3-540-76298-0\_52}

\bibitem{DBLP:conf/nips/BordesUGWY13}
Bordes, A., Usunier, N., Garc{\'{\i}}a{-}Dur{\'{a}}n, A., Weston, J.,
  Yakhnenko, O.: Translating embeddings for modeling multi-relational data. In:
  Burges, C.J.C., Bottou, L., Ghahramani, Z., Weinberger, K.Q. (eds.) Advances
  in Neural Information Processing Systems 26: 27th Annual Conference on Neural
  Information Processing Systems 2013. Proceedings of a meeting held December
  5-8, 2013, Lake Tahoe, Nevada, United States. pp. 2787--2795 (2013),
  \url{http://papers.nips.cc/paper/5071-translating-embeddings-for-modeling-multi-relational-data}

\bibitem{DBLP:conf/acl/CaoLLLLC19}
Cao, Y., Liu, Z., Li, C., Liu, Z., Li, J., Chua, T.: Multi-channel graph neural
  network for entity alignment. In: Korhonen et~al.
  \cite{DBLP:conf/acl/2019-1}, pp. 1452--1461,
  \url{https://www.aclweb.org/anthology/P19-1140/}

\bibitem{DBLP:conf/ijcai/ChenTCSZ18}
Chen, M., Tian, Y., Chang, K., Skiena, S., Zaniolo, C.: Co-training embeddings
  of knowledge graphs and entity descriptions for cross-lingual entity
  alignment. In: Lang  \cite{DBLP:conf/ijcai/2018}, pp. 3998--4004.
  \doi{10.24963/ijcai.2018/556}, \url{https://doi.org/10.24963/ijcai.2018/556}

\bibitem{DBLP:conf/ijcai/ChenTYZ17}
Chen, M., Tian, Y., Yang, M., Zaniolo, C.: Multilingual knowledge graph
  embeddings for cross-lingual knowledge alignment. In: Sierra
  \cite{DBLP:conf/ijcai/2017}, pp. 1511--1517. \doi{10.24963/ijcai.2017/209},
  \url{https://doi.org/10.24963/ijcai.2017/209}

\bibitem{anonymous2020deep}
Fey, M., Lenssen, J.E., Morris, C., Masci, J., Kriege, N.M.: Deep graph
  matching consensus. In: International Conference on Learning Representations
  (2020), \url{https://openreview.net/forum?id=HyeJf1HKvS}

\bibitem{gilmer2017neural}
Gilmer, J., Schoenholz, S.S., Riley, P.F., Vinyals, O., Dahl, G.E.: Neural
  message passing for quantum chemistry. In: Proceedings of the 34th
  International Conference on Machine Learning-Volume 70. pp. 1263--1272. JMLR.
  org (2017)

\bibitem{DBLP:conf/icml/GuoSH19}
Guo, L., Sun, Z., Hu, W.: Learning to exploit long-term relational dependencies
  in knowledge graphs. In: Chaudhuri, K., Salakhutdinov, R. (eds.) Proceedings
  of the 36th International Conference on Machine Learning, {ICML} 2019, 9-15
  June 2019, Long Beach, California, {USA}. Proceedings of Machine Learning
  Research, vol.~97, pp. 2505--2514. {PMLR} (2019),
  \url{http://proceedings.mlr.press/v97/guo19c.html}

\bibitem{kipf2016semi}
Kipf, T.N., Welling, M.: Semi-supervised classification with graph
  convolutional networks. arXiv preprint arXiv:1609.02907  (2016)

\bibitem{DBLP:conf/acl/2019-1}
Korhonen, A., Traum, D.R., M{\`{a}}rquez, L. (eds.): Proceedings of the 57th
  Conference of the Association for Computational Linguistics, {ACL} 2019,
  Florence, Italy, July 28- August 2, 2019, Volume 1: Long Papers. Association
  for Computational Linguistics (2019),
  \url{https://www.aclweb.org/anthology/volumes/P19-1/}

\bibitem{DBLP:conf/ijcai/2019}
Kraus, S. (ed.): Proceedings of the Twenty-Eighth International Joint
  Conference on Artificial Intelligence, {IJCAI} 2019, Macao, China, August
  10-16, 2019. ijcai.org (2019). \doi{10.24963/ijcai.2019},
  \url{https://doi.org/10.24963/ijcai.2019}

\bibitem{DBLP:conf/ijcai/2018}
Lang, J. (ed.): Proceedings of the Twenty-Seventh International Joint
  Conference on Artificial Intelligence, {IJCAI} 2018, July 13-19, 2018,
  Stockholm, Sweden. ijcai.org (2018),
  \url{http://www.ijcai.org/proceedings/2018/}

\bibitem{DBLP:conf/cidr/MahdisoltaniBS15}
Mahdisoltani, F., Biega, J., Suchanek, F.M.: {YAGO3:} {A} knowledge base from
  multilingual wikipedias. In: {CIDR} 2015, Seventh Biennial Conference on
  Innovative Data Systems Research, Asilomar, CA, USA, January 4-7, 2015,
  Online Proceedings. www.cidrdb.org (2015),
  \url{http://cidrdb.org/cidr2015/Papers/CIDR15\_Paper1.pdf}

\bibitem{nickel2015review}
Nickel, M., Murphy, K., Tresp, V., Gabrilovich, E.: A review of relational
  machine learning for knowledge graphs. Proceedings of the IEEE
  \textbf{104}(1),  11--33 (2015)

\bibitem{DBLP:conf/www/PeiYHZ19}
Pei, S., Yu, L., Hoehndorf, R., Zhang, X.: Semi-supervised entity alignment via
  knowledge graph embedding with awareness of degree difference. In: Liu, L.,
  White, R.W., Mantrach, A., Silvestri, F., McAuley, J.J., Baeza{-}Yates, R.,
  Zia, L. (eds.) The World Wide Web Conference, {WWW} 2019, San Francisco, CA,
  USA, May 13-17, 2019. pp. 3130--3136. {ACM} (2019).
  \doi{10.1145/3308558.3313646}, \url{https://doi.org/10.1145/3308558.3313646}

\bibitem{pei2019semi}
Pei, S., Yu, L., Hoehndorf, R., Zhang, X.: Semi-supervised entity alignment via
  knowledge graph embedding with awareness of degree difference. In: The World
  Wide Web Conference. pp. 3130--3136. ACM (2019)

\bibitem{DBLP:conf/ijcai/2017}
Sierra, C. (ed.): Proceedings of the Twenty-Sixth International Joint
  Conference on Artificial Intelligence, {IJCAI} 2017, Melbourne, Australia,
  August 19-25, 2017. ijcai.org (2017),
  \url{http://www.ijcai.org/Proceedings/2017/}

\bibitem{singhal2012introducing}
Singhal, A.: Introducing the knowledge graph: things, not strings. Official
  google blog  \textbf{5} (2012)

\bibitem{DBLP:conf/semweb/SunHL17}
Sun, Z., Hu, W., Li, C.: Cross-lingual entity alignment via joint
  attribute-preserving embedding. In: d'Amato, C., Fern{\'{a}}ndez, M., Tamma,
  V.A.M., L{\'{e}}cu{\'{e}}, F., Cudr{\'{e}}{-}Mauroux, P., Sequeda, J.F.,
  Lange, C., Heflin, J. (eds.) The Semantic Web - {ISWC} 2017 - 16th
  International Semantic Web Conference, Vienna, Austria, October 21-25, 2017,
  Proceedings, Part {I}. Lecture Notes in Computer Science, vol. 10587, pp.
  628--644. Springer (2017). \doi{10.1007/978-3-319-68288-4\_37},
  \url{https://doi.org/10.1007/978-3-319-68288-4\_37}

\bibitem{DBLP:conf/ijcai/SunHZQ18}
Sun, Z., Hu, W., Zhang, Q., Qu, Y.: Bootstrapping entity alignment with
  knowledge graph embedding. In: Lang  \cite{DBLP:conf/ijcai/2018}, pp.
  4396--4402. \doi{10.24963/ijcai.2018/611},
  \url{https://doi.org/10.24963/ijcai.2018/611}

\bibitem{sun2018bootstrapping}
Sun, Z., Hu, W., Zhang, Q., Qu, Y.: Bootstrapping entity alignment with
  knowledge graph embedding. In: IJCAI. pp. 4396--4402 (2018)

\bibitem{DBLP:conf/aaai/TrisedyaQZ19}
Trisedya, B.D., Qi, J., Zhang, R.: Entity alignment between knowledge graphs
  using attribute embeddings. In: The Thirty-Third {AAAI} Conference on
  Artificial Intelligence, {AAAI} 2019, The Thirty-First Innovative
  Applications of Artificial Intelligence Conference, {IAAI} 2019, The Ninth
  {AAAI} Symposium on Educational Advances in Artificial Intelligence, {EAAI}
  2019, Honolulu, Hawaii, USA, January 27 - February 1, 2019. pp. 297--304.
  {AAAI} Press (2019),
  \url{https://aaai.org/ojs/index.php/AAAI/article/view/3798}

\bibitem{velivckovic2017graph}
Veli{\v{c}}kovi{\'c}, P., Cucurull, G., Casanova, A., Romero, A., Lio, P.,
  Bengio, Y.: Graph attention networks. arXiv preprint arXiv:1710.10903  (2017)

\bibitem{DBLP:conf/emnlp/WangLLZ18}
Wang, Z., Lv, Q., Lan, X., Zhang, Y.: Cross-lingual knowledge graph alignment
  via graph convolutional networks. In: Riloff, E., Chiang, D., Hockenmaier,
  J., Tsujii, J. (eds.) Proceedings of the 2018 Conference on Empirical Methods
  in Natural Language Processing, Brussels, Belgium, October 31 - November 4,
  2018. pp. 349--357. Association for Computational Linguistics (2018),
  \url{https://www.aclweb.org/anthology/D18-1032/}

\bibitem{wikidata}
Wikidata. \url{https://www.wikidata.org/}

\bibitem{xu2019cross}
Xu, K., Wang, L., Yu, M., Feng, Y., Song, Y., Wang, Z., Yu, D.: Cross-lingual
  knowledge graph alignment via graph matching neural network. arXiv preprint
  arXiv:1905.11605  (2019)

\bibitem{DBLP:conf/acl/XuWYFSWY19}
Xu, K., Wang, L., Yu, M., Feng, Y., Song, Y., Wang, Z., Yu, D.: Cross-lingual
  knowledge graph alignment via graph matching neural network. In: Korhonen
  et~al.  \cite{DBLP:conf/acl/2019-1}, pp. 3156--3161,
  \url{https://www.aclweb.org/anthology/P19-1304/}

\bibitem{DBLP:conf/ijcai/ZhangSHCGQ19}
Zhang, Q., Sun, Z., Hu, W., Chen, M., Guo, L., Qu, Y.: Multi-view knowledge
  graph embedding for entity alignment. In: Kraus  \cite{DBLP:conf/ijcai/2019},
  pp. 5429--5435. \doi{10.24963/ijcai.2019/754},
  \url{https://doi.org/10.24963/ijcai.2019/754}

\bibitem{DBLP:conf/ijcai/ZhuXLS17}
Zhu, H., Xie, R., Liu, Z., Sun, M.: Iterative entity alignment via joint
  knowledge embeddings. In: Sierra  \cite{DBLP:conf/ijcai/2017}, pp.
  4258--4264. \doi{10.24963/ijcai.2017/595},
  \url{https://doi.org/10.24963/ijcai.2017/595}

\bibitem{DBLP:conf/ijcai/ZhuZ0TG19}
Zhu, Q., Zhou, X., Wu, J., Tan, J., Guo, L.: Neighborhood-aware attentional
  representation for multilingual knowledge graphs. In: Kraus
  \cite{DBLP:conf/ijcai/2019}, pp. 1943--1949. \doi{10.24963/ijcai.2019/269},
  \url{https://doi.org/10.24963/ijcai.2019/269}

\end{thebibliography}

\clearpage
\section*{Appendix}
\section*{Dataset Links}
\begin{center}
\begin{tabular}{*{2}{l}}
\toprule
Dataset & Link \\
\midrule
DBP15K-JAPE & \url{github.com/nju-websoft/JAPE/blob/master/data/dbp15k.tar.gz} \\
DBP15K-full & \url{ws.nju.edu.cn/jape/} \\
DWY100K & \url{github.com/nju-websoft/BootEA/tree/master/dataset/DWY100K} \\
WK3l-15K & \url{drive.google.com/open?id=1AsPPU4ka1Rc9u-XYMGWtvV65hF3egi0z} \\
WK3l-60k & \url{github.com/muhaochen/MTransE-tf/blob/master/preprocess/wk3l_60k.zip} \\
WK3l-120K & \url{drive.google.com/open?id=1AsPPU4ka1Rc9u-XYMGWtvV65hF3egi0z} \\
CN3l & \url{drive.google.com/open?id=1AsPPU4ka1Rc9u-XYMGWtvV65hF3egi0z}\\
DFB-1 & \url{github.com/thunlp/IEAJKE/tree/master/data}
\\
\bottomrule
\end{tabular}
\end{center}

\section*{Code Links}
\begin{center}
\begin{tabular*}{\linewidth}{*{4}{@{\extracolsep{\fill}}l}@{\extracolsep{\fill}}}
\toprule
Method & Link \\
\midrule
MTransE \cite{DBLP:conf/ijcai/ChenTYZ17} & \url{https://github.com/muhaochen/MTransE}\\
IPTransE \cite{DBLP:conf/ijcai/ZhuXLS17} & \url{https://github.com/thunlp/IEAJKE}\\
JAPE \cite{DBLP:conf/semweb/SunHL17} & \url{https://github.com/nju-websoft/JAPE}\\
KDCoE \cite{DBLP:conf/ijcai/ChenTCSZ18} & \url{https://github.com/muhaochen/MTransE-tf} \\
BootEA \cite{DBLP:conf/ijcai/SunHZQ18} & \url{https://github.com/nju-websoft/BootEA}\\
SEA \cite{DBLP:conf/www/PeiYHZ19} & \url{https://github.com/scpei/SEA} \\
MultiKE \cite{DBLP:conf/ijcai/ZhangSHCGQ19} & \url{https://github.com/nju-websoft/MultiKE} \\
AttrE \cite{DBLP:conf/aaai/TrisedyaQZ19} & \url{http://www.ruizhang.info/GKB/gkb.htm} \\
RSN \cite{DBLP:conf/icml/GuoSH19} & \url{https://github.com/nju-websoft/RSN}\\
GCN-Align \cite{DBLP:conf/emnlp/WangLLZ18} & \url{https://github.com/1049451037/GCN-Align}\\
CL-GNN \cite{DBLP:conf/acl/XuWYFSWY19} & \url{https://github.com/syxu828/Crosslingula-KG-Matching} \\
MuGNN \cite{DBLP:conf/acl/CaoLLLLC19} & \url{https://github.com/thunlp/MuGNN}\\
NAEA \cite{DBLP:conf/ijcai/ZhuZ0TG19} & - \\
\bottomrule
\end{tabular*}    
\end{center}
\begin{table}
    \centering
    \caption{Optimal Hyperparameters after finetuning LR, number of epochs and number of layers for each individual dataset / subset combination. We only report differences to the ones found on DBP15k (JAPE) zh-en.}
    \begin{tabular*}{\linewidth}{*{4}{l}*{3}{@{\extracolsep{\fill}}r}@{\extracolsep{\fill}}}
        \toprule
        var. emb. init & weights & dataset & subset&  \#epochs &  \#layers &    lr \\
        \midrule
        1 & no & DBP15k (full) & fr-en &         2k &         2 &   1.0 \\
                 &     &         & ja-en &         2k &         3 &   1.0 \\
                 &     &         & zh-en &         2k &         4 &   1.0 \\
                 &     & DBP15k (JAPE) & fr-en &         2k &         2 &   1.0 \\
                 &     &         & ja-en &         2k &         2 &   1.0 \\
                 &     &         & zh-en &         2k &         2 &   1.0 \\
                 &     & DWY100k & wd &         2k &         2 &   1.0 \\
                 &     &         & yg &         2k &         2 &   1.0 \\
                 &     & WK3l-120k & en-de &         2k &         2 &   1.0 \\
                 &     &         & en-fr &         2k &         2 &   1.0 \\
                 &     & WK3l-15k & en-de &         2k &         2 &   1.0 \\
                 &     &         & en-fr &         2k &         2 &  10.0 \\
                 & yes & DBP15k (full) & fr-en &         2k &         4 &   1.0 \\
                 &     &         & ja-en &         2k &         4 &   1.0 \\
                 &     &         & zh-en &         2k &         3 &   1.0 \\
                 &     & DBP15k (JAPE) & fr-en &         2k &         2 &  10.0 \\
                 &     &         & ja-en &         2k &         3 &   1.0 \\
                 &     &         & zh-en &         2k &         3 &   1.0 \\
                 &     & DWY100k & wd &         2k &         2 &   1.0 \\
                 &     &         & yg &         2k &         2 &   1.0 \\
                 &     & WK3l-120k & en-de &         2k &         2 &   1.0 \\
                 &     &         & en-fr &         2k &         2 &   1.0 \\
                 &     & WK3l-15k & en-de &         2k &         2 &   1.0 \\
                 &     &         & en-fr &         2k &         2 &   1.0 \\
        $n^{-1/2}$ & no & DBP15k (full) & fr-en &         3k &         2 &   1.0 \\
                 &     &         & ja-en &         3k &         2 &   1.0 \\
                 &     &         & zh-en &         2k &         4 &   1.0 \\
                 &     & DBP15k (JAPE) & fr-en &         3k &         2 &   1.0 \\
                 &     &         & ja-en &         2k &         2 &   1.0 \\
                 &     &         & zh-en &         3k &         2 &   1.0 \\
                 &     & DWY100k & wd &         3k &         2 &   1.0 \\
                 &     &         & yg &         3k &         2 &   1.0 \\
                 &     & WK3l-120k & en-de &         3k &         2 &   0.5 \\
                 &     &         & en-fr &         3k &         2 &   1.0 \\
                 &     & WK3l-15k & en-de &         3k &         2 &   0.5 \\
                 &     &         & en-fr &         3k &         2 &   1.0 \\
                 & yes & DBP15k (full) & fr-en &         2k &         4 &   1.0 \\
                 &     &         & ja-en &         2k &         4 &   1.0 \\
                 &     &         & zh-en &         2k &         4 &   1.0 \\
                 &     & DBP15k (JAPE) & fr-en &         2k &         2 &   1.0 \\
                 &     &         & ja-en &         2k &         2 &   1.0 \\
                 &     &         & zh-en &         2k &         2 &   1.0 \\
                 &     & DWY100k & wd &         2k &         2 &   1.0 \\
                 &     &         & yg &         3k &         2 &   0.5 \\
                 &     & WK3l-120k & en-de &         2k &         2 &   1.0 \\
                 &     &         & en-fr &         2k &         2 &   1.0 \\
                 &     & WK3l-15k & en-de &         2k &         2 &   1.0 \\
                 &     &         & en-fr &         2k &         2 &   1.0 \\
        \bottomrule
    \end{tabular*}
    \label{tab:best_hparams}
\end{table} 
\end{document}